%% file: main.tex
\documentclass[sigconf]{acmart}

\AtBeginDocument{%
  \providecommand\BibTeX{{%
    \normalfont B\kern-0.5em{\scshape i\kern-0.25em b}\kern-0.8em\TeX}}}

\copyrightyear{2024}
\acmYear{2024}
\setcopyright{acmlicensed}\acmConference[WSDM '24]{Proceedings of the 17th ACM International Conference on Web Search and Data Mining}{March 4--8, 2024}{Merida, Mexico}
\acmBooktitle{Proceedings of the 17th ACM International Conference on Web Search and Data Mining (WSDM '24), March 4--8, 2024, Merida, Mexico}
\acmPrice{15.00}
\acmDOI{10.1145/3616855.3635844}
\acmISBN{979-8-4007-0371-3/24/03}

\include{commands}
\usepackage{amsmath}
\usepackage{multirow}
\usepackage{enumitem}
\usepackage{bbm}
\usepackage{xcolor}
\DeclareMathOperator*{\argmax}{arg\,max}
\DeclareMathOperator*{\argmin}{arg\,min}
\newcommand*{\Scale}[2][4]{\scalebox{#1}{$#2$}}%

\newcommand{\wsdmnew}[1]{{\textcolor{black}{#1}}}
\begin{document}
\settopmatter{printacmref=false}

\title{FairIF: Boosting Fairness in Deep Learning via Influence Functions with Validation Set Sensitive Attributes}

\author{Haonan Wang}
\authornote{Both authors contributed equally to this research.}
\email{haonan.wang@u.nus.edu}
\affiliation{%
  \institution{National University of Singapore}
  \country{Singapore}
}

\author{Ziwei Wu}
\authornotemark[1]
\email{ziweiwu2@illinois.edu}
\affiliation{%
  \institution{University of Illinois at Urbana-Champaign}
  \country{USA}
}

\author{Jingrui He}
\email{jingrui@illinois.edu}
\affiliation{%
  \institution{University of Illinois at Urbana-Champaign}
  \country{USA}
}

\input{0_abstract}

\ccsdesc[500]{Computing methodologies~Machine learning algorithms}


\maketitle

\input{1_intro}
\input{2_related}
\input{3_prelim}
\input{4_method}
\input{5_analysis}
\input{6_experiments}

\input{7_conclusion}

\bibliographystyle{ACM-Reference-Format}
\bibliography{ref}
\newpage
\input{8_appendix}
\end{document}

%% file: commands.tex
\usepackage{booktabs} 
\usepackage{amsfonts}
\usepackage{amsthm}
\usepackage{ctable}
\usepackage{multirow}
\usepackage{pifont}
\usepackage{color}
\usepackage{subfigure}
\usepackage{enumitem}
\usepackage{sidecap}
\usepackage{xcolor}
\usepackage[normalem]{ulem}

\usepackage[ruled,vlined]{algorithm2e}
\usepackage{algorithmic,amsmath}
\usepackage{bm}

\newcommand{\rom}[1]{\uppercase\expandafter{\romannumeral #1\relax}}

\newcommand{\mylistbegin}{
  \begin{list}{$\bullet$}
   {
     \setlength{\itemsep}{-2pt}
     \setlength{\leftmargin}{1em}
     \setlength{\labelwidth}{1em}
     \setlength{\labelsep}{0.5em} } }
\newcommand{\mylistend}{
   \end{list}  }

\newcommand{\eg}{\textit{e.g.,}}

\newcommand{\ie}{\textit{i.e.}}

\newcommand{\header}[1]{{\vspace{+1mm}\flushleft \textbf{#1}}}

\newtheorem{prop}{Proposition}

\newtheorem{definition}{Definition}

\newtheorem{thm:def}{Definition}
\newtheorem{thm:lemma}{Lemma}

\newcommand{\fairIF}{\textsc{FairIF}}

\makeatletter
\newcommand{\nosemic}{\renewcommand{\@endalgocfline}{\relax}}

%% file: 0_abstract.tex
\begin{abstract}
Empirical loss minimization during machine learning training can inadvertently introduce bias, stemming from discrimination and societal prejudices present in the data. To address the shortcomings of traditional fair machine learning methods—which often rely on sensitive information of training data or mandate significant model alterations—we present \fairIF, a unique two-stage training framework. Distinctly, \fairIF\ enhances fairness by recalibrating training sample weights using the influence function. Notably, it employs sensitive information from a validation set, rather than the training set, to determine these weights. This approach accommodates situations with missing or inaccessible sensitive training data. Our \fairIF\ ensures fairness across demographic groups by retraining models on the reweighted data. It stands out by offering a plug-and-play solution, obviating the need for changes in model architecture or the loss function.
We demonstrate that the fairness performance of \fairIF~ is guaranteed during testing with only a minimal impact on classification performance. Additionally, we analyze that our framework adeptly addresses issues like group size disparities, distribution shifts, and class size discrepancies.
Empirical evaluations on three synthetic and five real-world datasets across six model architectures confirm \fairIF's efficiency and scalability. The experimental results indicate superior fairness-utility trade-offs compared to other methods, regardless of bias types or architectural variations. Moreover, the adaptability of \fairIF\ to utilize pretrained models for subsequent tasks and its capability to rectify unfairness originating during the pretraining phase are further validated through our experiments.
\end{abstract}

%% file: 1_intro.tex
\section{Introduction}
\label{sec:intro}

In automatic, high-stake decision-making systems, the use of machine learning techniques is commonplace. In spite of the effectiveness of these machine learning techniques, recent works~\cite{hajian2016algorithmic} have uncovered algorithmic discrimination across demographic groups in real-world applications, which raises severe fairness concerns~\cite{ekstrand2018all}. In response, there has been a flurry of research on fairness in machine learning~\cite{mehrabi2021survey, caton2020fairness}, with a primary emphasis on proposing formal notions of fairness~\cite{hardt2016equality,zafar2017fairness} and ``de-biasing'' techniques to achieve these goals~\cite{agarwal2018reductions,feldman2015certifying}.

The issue of indirect discrimination in deep learning algorithms has garnered substantial research interest because of its profound implications for output fairness. Addressing it directly, such as removing sensitive attributes during training, is inadequate to guarantee equality due to its intricate causes~\cite{castelnovo2022clarification}.
In order to ensure that the system is not biased against some sensitive features, previous methods either heavily rely on the sensitive information in the training data to construct training objective functions~\cite{dwork2012fairness,zafar2017faircons}, or add additional modules
 to the original model to ensure fairness and balance in the predictions despite the presence of potentially biased data~\cite{adel2019one,zhang2018mitigating}.
In general, the vast majority of works on fairness assume that sensitive information, such as gender or race, is contained in the training set and that the model's design is completely accessible and modifiable~\cite{lahoti2020fairness,madras2018learning}. 
However, in many scenarios, it is difficult to gather or use sensitive information for decision-making due to privacy or legal restrictions~\cite{voigt2017eu}. 
Moreover, in numerous real-world applications~\cite{wang2021deep}, the developed models rigorously adhere to state-of-the-art architectures to achieve the desired performance. Modifying complex designs, \eg~ introducing an additional Multi-Layer Perceptron (MLP) branch for fairness, will degrade the performance or introduce extra complication into the model tuning. 
Therefore, algorithms that rely solely on sensitive data or require substantial modifications to target methods are typically difficult to implement in practice. 
In this study, we pose the following research question:

\vspace{3pt}
\textbf{\textit{ 
Can we design a practical method which can yield models with better fairness performance when we cannot modify the architecture of the target model or have access to a great amount of sensitive information?
}}
\vspace{3pt}


We provide an affirmative answer to the above question and introduce the \fairIF\ model, a mechanism that enhances fairness in models by reweighting training samples.
Distinctively, \fairIF\ preserves the model's original architecture, facilitating its seamless integration with a broad spectrum of machine learning models optimized by gradient descent.  This framework promotes enhanced fairness metrics including equality of opportunity, odds, and accuracy. Crucially, \fairIF\ doesn't mandate the inclusion of sensitive data within the training samples—only a modest validation set annotated with group labels is needed. The methodology unfolds in two phases. 
Initially, the Influence Function (IF)~\cite{koh2017understanding} is employed to quantify sample influence, \ie, the change in model prediction if a training sample is up- (down-)weighted by a specific amount. Subsequently, sample weights are optimized to ensure that the given model achieves performance equality across all groups of the validation set. In the succeeding phase, the model undergoes retraining on the weighted dataset, addressing fairness concerns while maintaining inherent performance. Furthermore, an analysis underscores that the performance gap between the original and retrained models remains constrained, and the fairness performance is guaranteed when tested.

The experiments are threefold. First, on synthetic CI-MNIST dataset~\cite{reddy2021benchmarking}, by independently generating three different types of bias, (1) difference group sizes, (2) difference class distributions within each group, and (3) difference class sizes, \fairIF\ successfully mitigates the unfairness caused by different group sizes, class distributions, and class sizes. This shows that the method can handle a variety of biases in a controlled environment.
Second, with commonly used deep neural networks on three real-world fairness datasets and two image datasets~\cite{liu2015deep,karkkainen2019fairface},
\fairIF\ improves the values of multiple fairness metrics while maintaining or even enhancing model accuracy. This performance is superior to five previous state-of-the-art methods. This illustrates that \fairIF\ not only scales well but also surpasses other methods in handling complex, uncontrolled, real-world data.
Third, even when used with different pretrained models, \fairIF\ manages to alleviate the unfairness within these models without detriment to their performance. This highlights the adaptability and versatility of \fairIF\ across different model architectures and initializations.
Additionally, we empirically analyze the role of the validation set in \fairIF. The result indicates that \fairIF\ is able to achieve desired performance with only a small amount of validation set, which makes the proposed method feasible and desirable for real-world applications in which sensitive information is hard to collect.
Besides, to further study how our method achieves fairness, we also examine what examples are reweighted by \fairIF.  Examining the specific examples reweighted by \fairIF\ provides one way to better understand the intermediate process, which is critical for trustworthiness and further improvement of the method.

%% file: 2_related.tex
\vspace{-8mm}
\section{Related Work}
\textbf{Influence Function.}
Originating in 1970s statistics, the influence function was designed to measure a model's dependence on specific training samples. It was later incorporated into machine learning, notably aiding in interpreting predictions by assessing each sample's impact~\cite{koh2017understanding}. Numerous extensions followed, such as Barshan et al.'s work focusing on local influence relative to global effects~\cite{barshan2020relatif}, and Basu et al.'s exploration of the collective influence of large training groups~\cite{basu2020second, koh2019accuracy}. Other studies accelerated inference for overparameterized neural networks~\cite{borsos2020coresets, guo2020fastif}. Of note, existing influence function methods assume the Hessian matrix's positive definiteness. Many dynamically compute this before model convergence, risking inaccuracies. Our \fairIF\ method awaits full model convergence, ensuring more accurate and time-efficient computations.\\
\textbf{Fairness-aware machine learning.}
Fair machine learning aims to counteract biases in automated systems. In classification, group fairness demands consistent classification error across protected-attribute groups~\cite{zafar2017fairness, hardt2016equality}. Existing solutions often involve extensive model and training modifications~\cite{zhao2019conditional,zhang2018mitigating,wu2022fairness} or rely on sensitive data~\cite{grgic2018beyond,wang2022understanding}. Some address cases without the sensitive attribute~\cite{yan2020fair, lahoti2020fairness} while others use sample reweighing~\cite{krasanakis2018adaptive, jiang2020identifying}, which can be costly due to frequent retraining. From model repair, methods have been proposed leveraging counterfactual distributions~\cite{wang2018avoiding} or sample influence estimation~\cite{sattigeri2022fair}. While some, like Wang et al., employed the Influence Function for instance-level fairness constraints~\cite{wang2022understanding}, others used it to compute sample weights to bridge fairness gaps~\cite{li2022achieving}. However, such methods often need entire training sets' sensitive data or entail computationally demanding steps, such as computing the Hessian matrix per sample and using a solver for linear programming problems, making them unsuitable for large-scale datasets in deep learning models~\cite{li2022achieving, wang2022understanding}.
Our \fairIF\ method stands out, ensuring fairness using only group annotations on a small validation set, without changing the model's architecture. This introduces a fresh avenue for bolstering model fairness.

%% file: 3_prelim.tex
\section{Preliminary}
\label{sec:pre}
\subsection{Notation}
\label{sec:notation}
 We consider the setting where each sample consists of an input $x \in \mathcal{X}$, a label $y \in \mathcal{Y}$, where $\mathcal{X}$ and $\mathcal{Y}$ are the input and output space respectively, and each example has a corresponding sensitive attribute $s \in \mathcal{S}$. For simplicity’s sake, assume that $\mathcal{S} = \{0,1\}$.  Let $K$ denotes the number of classes, $[K] := \{1,2,..,K\}$. We mainly focus on the classification problem and denote the classifier $h(x) = \argmax_{i \in [K]} f^{i}_{\theta}(x)$, where $f_{\theta} \in \mathbb{R}^K$ is a neural network parameterized by $\theta \in \Theta$. We denote the number of parameters as $P$, then  $\Theta \subseteq \mathbb{R}^P$.
Denote the training data set of size $n$ as $\mathcal{D} = \{z_1, z_2, ..., z_n\}$, where $z_i = (x_i, y_i) \in \mathcal{X} \times \mathcal{Y}$, and the validation data set (with sensitive attributes) of size $m$ as $\mathcal{D}^s = \{z^s_1, z^s_2, ..., z^s_m\}$, where $z^s_j = (x_j, y_j, s_j) \in \mathcal{X} \times \mathcal{Y} \times \mathcal{S}$.
In this work, we are interested in the setting where the sensitive attribute $s$ is not available for the training set, as collecting large amount of data with the sensitive attributes is typically expensive and at the risk of privacy leakage \cite{yan2020fair}. Instead, we assume that we have access to a small validation set with annotations of the sensitive attribute. 
The standard training procedure 
minimizes the empirical risk ${\mathcal{L}}(\mathcal{D},\theta)$,
\begin{equation}
\begin{small}
\begin{aligned}
\theta^{\star} = \argmin_{\theta \in \Theta}{\mathcal{L}}(\mathcal{D},\theta) = \sum_{i=1}^n \frac{1}{n}~\ell\left(z_i, \theta\right),
\end{aligned}
\end{small}
\end{equation}
where $\ell\left(z_i, \theta\right):\mathcal{X} \times \mathcal{Y} \times \Theta \rightarrow \mathbb{R}^{+}$ is the loss function (\eg cross entropy loss function) for $z_i$. 
Our goal is to learn a fair neural network $f_{\theta^{\star}_{\boldsymbol{\epsilon}}}$ parameterized by $\theta^{\star}_{\boldsymbol{\epsilon}} \in \Theta$, through minimizing the weighted loss ${\mathcal{L}}_{\boldsymbol{\epsilon}}$:
\vspace{-1mm}
\begin{equation}
\begin{small}
\begin{aligned}
\theta^{\star}_{\boldsymbol{\epsilon}} = \argmin_{\theta \in \Theta} {\mathcal{L}}_{\boldsymbol{\epsilon}}(\mathcal{D},\theta) = \argmin_{\theta \in \Theta} \sum_{i=1}^n (\frac{1}{n} + \epsilon_i)~\ell\left(z_i, \theta\right),
\end{aligned}
\end{small}
\vspace{-1mm}
\end{equation}
where $\boldsymbol{\epsilon} = [\epsilon_1, \epsilon_2, ..., \epsilon_n]^{\top} \in \mathbb{R}^n$ is a reweight vector of the training samples. 
Next, we introduce the fairness notions used in this work.
For notation convenience, we denote the true positive rate on group 1 (sensitive attribute $s=1$) as $\text{TPR}^{(1)} = \mathbb{P}(h=1 \mid s=1, y=1)$ where $h$ is the classifer's prediction, and the true positive rate on group 0 (sensitive attribute $s=0$) as $\text{TPR}^{(0)} = \mathbb{P}(h=1 \mid s=0, y=1)$. Similarly, we denote the true negative rate on group 1 as $\text{TNR}^{(1)} = \mathbb{P}(h=0 \mid s=1, y=0)$ and the true negative rate on group 0 as $\text{TNR}^{(0)} = \mathbb{P}(h=0 \mid s=0, y=0)$.

\header{Accuracy Equality} requires the classification system to have equal misclassification rates across sensitive groups~\cite{zafar2017fairness}:  $\mathbb{P}(h(x) = y | s=0) = \mathbb{P}(h(x) = y | s=1) $.
Then, we define the {\it Accuracy Difference (AD)} as:
\vspace{-1mm}
\begin{equation}
\label{equ:AD}
\begin{small}
\begin{aligned}
AD = | \mathbb{P}(h(x) = y | s=0) - \mathbb{P}(h(x) = y | s=1) |.
\end{aligned}
\end{small}
\vspace{-2mm}
\end{equation}

\header{Equal Odds} is defined as $\mathbb{P}(h=1 \mid s=1, y=\wsdmnew{y^\prime}) = \mathbb{P}(h=1 \mid s=0, y=\wsdmnew{y^\prime), \forall y^\prime \in \{0,1\}}$. It is sometimes also referred to as disparate mistreatment, aiming to equalize the {\it true positive} and {\it false positive rates} for a (binary-) classifier~\cite{hardt2016equality}.
Following~\cite{ozdayi2021bifair,yan2020fair}, we define {\it Average Odds Difference (AOD)} as:
\vspace{-1mm}
\begin{equation}
\label{equ:AOD}
\begin{small}
\begin{aligned}
AOD=\frac{1}{2}\big[|\text{TPR}^{(1)} - \text{TPR}^{(0)}| + |\text{TNR}^{(1)} - \text{TNR}^{(0)}| \big].
\end{aligned}
\end{small}
\vspace{-2mm}
\end{equation}

\header{Equal Opportunity} is weaker than Equal Odds, but it typically allows for stronger utility~\cite{hardt2016equality}:
$\mathbb{P}(h=1 \mid s=1, y=1) = \mathbb{P}(h=1 \mid s=0, y=1)$.
Also, following~\cite{ozdayi2021bifair,yan2020fair}, we define {\it Equality of Opportunity Difference (EOD)} as:
\vspace{-1mm}
\begin{equation}
\label{equ:EOD}
\begin{small}
\begin{aligned}
EOD = |\text{TPR}^{(1)} - \text{TPR}^{(0)}|.
\end{aligned}
\end{small}
\vspace{-1mm}
\end{equation}

\subsection{Influence Function}
\label{sec:IF}
The method of Influence Function (IF)~\cite{koh2017understanding} aims at approximating how the minimizer of the loss function $\theta^{\star}$ would change if we were to reweight the $i$-th training example. The key idea is to make a first-order approximation of change in $\theta^{\star}$ around $\epsilon_i = 0$ with Taylor expansion. Specifically, if the $i$-th training sample is upweighted by a small $\epsilon_{i}$, then the perturbed risk minimizer $\theta^{\star}_{\epsilon_{i}}$ becomes:
\begin{equation}
\begin{aligned}
\theta^{\star}_{\epsilon_{i}} \triangleq \argmin _{\theta \in \Theta} \frac{1}{n} \sum_{i=1}^{n} \ell\left(z_i, \theta\right)+\epsilon_{i}~\ell\left(z_i, \theta\right).
\end{aligned}
\end{equation}
The change of model parameters due to the introduction of the weight ${\epsilon_{i}}$ is:
\begin{equation}
\begin{aligned}
\label{equ:param_if}
 {\theta}^{\star}_{\epsilon_{i}} - {\theta}^{\star} \approx \frac{d {\theta}^{\star}_{\epsilon_{i}}}{d \epsilon_{i}}\Big|_{\epsilon_{i}=0} \epsilon_{i} =\underline{-H_{{\theta}^{\star}}^{-1} \nabla_{\theta} \ell(z_i, {\theta}^{\star} }) \epsilon_{i} = \underline{\mathcal{I}_{param}(z_i}) \epsilon_{i},
\end{aligned}
\end{equation}
where $H_{\theta^{\star}}=\frac{1}{n} \sum_{i=1}^n \nabla_{\theta}^{2} \ell\left(z_{i}, \theta^{\star}\right)$ is the Hessian of the objective at $\theta^{\star}$, and $\mathcal{I}_{param}(z_i) \in \mathbb{R}^P$ denotes the influence of sample $z_i$ on the model parameters.
Following the assumption adopted in previous works~\cite{koh2017understanding,koh2019accuracy,barshan2020relatif,teso2021interactive}, $H_{\theta^{\star}}$ is positive definite. Note, this assumption is relatively weak, under the condition that $\theta^{\star}$ is the minimizer of the loss function. Similarly, the change in loss can be approximated~\cite{barshan2020relatif} as:
\begin{equation}
\begin{small}
\label{equ:if_loss_change}
\begin{aligned}
\ell\left(z, \theta_{\epsilon_{i}}^{\star}\right)-\ell \left(z, \theta^{\star}\right)  & \approx \frac{{d} \ell\left(z, \theta_{\epsilon_{i}}^{\star}\right)}{{d} \epsilon_{i}} \epsilon_{i}\\
&=\underline{-\nabla_{\theta} \ell\left(z, \theta^{\star}\right)^{\top} H_{{\theta}^{\star}}^{-1} \nabla_{\theta} \ell(z_i, {\theta}^{\star}}) \epsilon_{i}\\
&= \underline{\mathcal{I}_{loss}(z_i, z}) \epsilon_{i},
\end{aligned}
\end{small}
\end{equation}
where the term $\mathcal{I}_{loss}(z_i, z) \in \mathbb{R}$ represents the influence of sample $z_i$ on the loss computed over sample $z$.

%% file: 4_method.tex
\section{Method}
\label{sec:method}
\subsection{Correcting Discrepancy by Sample Reweighting}
The foundational insight underpinning \fairIF~ is that the influence of a training sample on the model prediction can be estimated through the influence function. This approach has been theoretically proven to offer precise estimations for linear models and has been empirically validated as effective in real-world applications~\cite{koh2017understanding,koh2019accuracy,zhang2022understanding,grosse2023studying}. 
Specifically, For any continuously differentiable functions $F(\mathcal{D}, \theta)\in \mathbb{R}$, $\mathcal{D} \subseteq \mathcal{X} \times \mathcal{Y} $, the change of $F$ with respect to $\epsilon_i$, the sample weight of $z_i$, can be computed based on the influence function:
\begin{equation}
\begin{small}
\label{equ:F_diff_change}
\begin{aligned}
& F(\mathcal{D}, \theta^\star_{\epsilon_i}) - F(\mathcal{D}, \theta^\star) \\
& \approx \frac{1}{|\mathcal{D}|}\frac{d \sum_{j}^{|\mathcal{D}|} \big(F(\{z_{j}\}, \theta^\star_{\epsilon_i}) - F(\{z_{j}\}, \theta^\star)\big)}{d \theta^\star_{{\epsilon_i}}} \frac{d \theta^\star_{{\epsilon_i}}}{d \epsilon_i} \epsilon_i \\
& = - \frac{1}{|\mathcal{D}|}\sum_{j}^{|\mathcal{D}|} \nabla_\theta F(\{z_{j}\}, \theta^\star)^{\top} H^{-1}_{\theta^\star}\nabla_\theta \ell(z_{i}, \theta^\star) \epsilon_i.
\end{aligned}
\end{small}
\end{equation}
The Equation~\eqref{equ:F_diff_change} can be regarded as an extension of the conclusion of Equation~\eqref{equ:if_loss_change}.
To apply the conclusion of Equation~\eqref{equ:F_diff_change} to the fairness setting, we define the function $F$ as a general metric measuring fairness. Denote the two different groups of the validation set $\mathcal{D}^s$ as $\mathcal{D}^0$ and $\mathcal{D}^1$ respectively, where for any $z_j^s = (x_j, y_j, s_j) \in \mathcal{D}^0, s_j=0$ and $z_{j'} = (x_{j'}, y_{j'}, s_{j'}) \in \mathcal{D}^1, s_{j'}=1$.
Then, for the two groups, the change of function $F$ caused by permuting sample weights are,
\begin{equation}
\label{equ:d0}
\begin{footnotesize}
\begin{aligned}
&{F(\mathcal{D}^0, \theta^\star_{\mathbf{\epsilon}}) - F(\mathcal{D}^0, \theta^\star) = - \frac{1}{|\mathcal{D}^0|} \sum_{z_i \in \mathcal{D}, z_j \in \mathcal{D}^0} \nabla_\theta F(z_j)^{\top} H^{-1}_{\theta^\star} \nabla_\theta \ell(z_i, \wsdmnew{\theta^\star})  \epsilon_i,}
\end{aligned}
\end{footnotesize}
\end{equation}
and,
\begin{equation}
\label{equ:d1}
\begin{footnotesize}
\begin{aligned}
&{F(\mathcal{D}^1, \theta^\star_{\mathbf{\epsilon}}) - F(\mathcal{D}^1, \theta^\star) = - \frac{1}{|\mathcal{D}^1|} \sum_{z_i \in \mathcal{D}, z_{j'} \in \mathcal{D}^1} \nabla_\theta F(z_{j'})^{\top} H^{-1}_{\theta^\star}  \nabla_\theta \ell(z_i, \wsdmnew{\theta^\star})  \epsilon_i.}
\end{aligned}
\end{footnotesize}
\end{equation}

Then, the goal of achieving \wsdmnew{fairness is to achieve equalized performance $F$ over the two groups, i.e.,}  solving $\mathbf{\epsilon}^{\star}$ to satisfy the following equation,
\begin{equation}
\begin{footnotesize}
\label{equ:new_goal}
\begin{aligned}
F(\mathcal{D}^0, \theta^\star_{\mathbf{\epsilon}^{\star}}) - F(\mathcal{D}^1, \theta^\star_{\mathbf{\epsilon}^{\star}}) = 0.
\end{aligned}
\end{footnotesize}
\end{equation}
\wsdmnew{Combining Equations (\ref{equ:d0})(\ref{equ:d1})(\ref{equ:new_goal}), we have:}
\begin{equation*}
\begin{footnotesize}
\begin{aligned}
& F(\mathcal{D}^0, \theta^\star_{\mathbf{\epsilon^{\star}}}) - F(\mathcal{D}^1, \theta^\star_{\mathbf{\epsilon^{\star}}}) 
- \big(F(\mathcal{D}^0, \theta^\star) - F(\mathcal{D}^1, \theta^\star) \big) \\
& = \Big(\frac{1}{|\mathcal{D}^1|} \sum_{ z_{j'} \in \mathcal{D}^1} \nabla_\theta F(z_{j'})   - \frac{1}{|\mathcal{D}^0|} \sum_{z_j \in \mathcal{D}^0} 
 \nabla_\theta F(z_j) \Big)^{\top} H^{-1}_{\theta^\star} \sum_{z_i \in \mathcal{D}} \nabla_\theta \ell(z_i, \wsdmnew{\theta^\star})  \epsilon^{\star}_i,
\end{aligned}
\end{footnotesize}
\end{equation*}
Notably, the second term $F(\mathcal{D}^0, \theta^\star) - F(\mathcal{D}^1, \theta^\star)$ denotes the metric discrepancy of \wsdmnew{unweighted} model across two demographic groups. We denote this empirically measurable discrepancy under metric $F$ as $\text{diff}(\mathcal{D}^s, F, \theta^\star)$. 
Then we have the following equation,
\begin{equation*}
\begin{footnotesize}
\begin{aligned}
F(\mathcal{D}^0, \theta^\star_{\mathbf{\epsilon^{\star}}}) - F(\mathcal{D}^1, &\theta^\star_{\mathbf{\epsilon^{\star}}}) = \text{diff}(\mathcal{D}^s, F, \theta^\star) + \Big(\frac{1}{|\mathcal{D}^1|} \sum_{ z_{j'} \in \mathcal{D}^1} \nabla_\theta F(z_{j'})  \\
&   - \frac{1}{|\mathcal{D}^0|} \sum_{z_j \in \mathcal{D}^0} 
 \nabla_\theta F(z_j) \Big)^{\top} H^{-1}_{\theta^\star} \sum_{z_i \in \mathcal{D}} \nabla_\theta \ell(z_i,\wsdmnew{\theta^\star})  \epsilon^{\star}_i,
\end{aligned}
\end{footnotesize}
\end{equation*}
In order to find $\mathbf{\epsilon}^\star$ making $F(\mathcal{D}^0, \theta^\star_{\mathbf{\epsilon}^{\star}}) - F(\mathcal{D}^1, \theta^\star_{\mathbf{\epsilon}^{\star}}) = 0$, we introduce the following optimization problem,
\vspace{-0.1cm}
\begin{equation}
\begin{footnotesize}
\label{equ:eps_min_loss}
\begin{aligned}
\epsilon^\star= \argmin_{\epsilon} &  \Big[\text{diff}(\mathcal{D}^s, F, \theta^\star) + \Big(\frac{1}{|\mathcal{D}^1|} \sum_{ z_{j'} \in \mathcal{D}^1} \nabla_\theta F(z_{j'}) \\
& - \frac{1}{|\mathcal{D}^0|} \sum_{z_j \in \mathcal{D}^0} \nabla_\theta F(z_j) \Big)^{\top} H^{-1}_{\theta^\star} \sum_{z_i \in \mathcal{D}} \nabla_\theta \ell(z_i,\wsdmnew{\theta^\star})  \wsdmnew{\epsilon_i} \Big]^2.
\end{aligned}
\end{footnotesize}
\end{equation}
In the subsequent section, we will leverage this conclusion to address practical fairness challenges.

\subsection{\fairIF: Achieving Fairness through Influence Function Reweighting}
We now present \fairIF, a straightforward two-stage methodology that eliminates the need for group annotations within the training set or alterations to the original models. Initially, we determine the sample weights using the influence function to ensure a balanced TPR and TNR performance across varied groups. Subsequently, in the second stage, we train the final model utilizing the reweighted training samples.

\header{Stage One.} To address discrepancies across three fairness notions— AD, AOD, and AOE—\fairIF~ aims to equalize True Positive Rate (TPR) and True Negative Rate (TNR) between two groups by adjusting the sample weights, denoted as $\epsilon$. Section~\ref{sec:disparity_analysis} offers an in-depth analysis of metric selection, highlighting that equalizing TPR and TNR serves as an effective fairness objective and can alleviate disparities under the aforementioned notions. However, since TPR and TNR are non-differentiable, they impede the use of Equation~\eqref{equ:eps_min_loss} to determine $\epsilon$. To circumvent this, we adopt the gumbel softmax technique~\cite{jang2016categorical} to approximate and render TPR and TNR differentiable. These approximations are represented as $F_{TPR}$ and $F_{TNR}$ respectively.
Then, the objective can be defined as:
\begin{equation}
\label{equ:final_eps_loss}
\begin{footnotesize}
\begin{aligned}
\wsdmnew{\min_{\bm \epsilon}} & \Big[ \text{diff} (\mathcal{D}^s, F_{TPR}, \theta^\star) + \big(\frac{1}{|\mathcal{D}^1|} \sum_{ z_{j'} \in \mathcal{D}^1} \nabla_\theta{F_{TPR}}(z_{j'})   \\ 
&- \frac{1}{|\mathcal{D}^0|} \sum_{z_j \in \mathcal{D}^0} \nabla_\theta{F_{TPR}}(z_j)\big)^{\top} H^{-1}_{\theta^\star} \sum_{z_i \in \mathcal{D}} \nabla_\theta \ell(z_i, \wsdmnew{\theta^\star})  \epsilon_i \Big]^2   \\
&+ \Big[\text{diff} (\mathcal{D}^s, F_{TNR}, \theta^\star) + \big(\frac{1}{|\mathcal{D}^1|} \sum_{ z_{j'} \in \mathcal{D}^1} \nabla_\theta{F_{TNR}}(z_{j'})  \\
&- \frac{1}{|\mathcal{D}^0|} \sum_{z_j \in \mathcal{D}^0} \nabla_\theta{F_{TNR}}(z_j)\big)^{\top} H^{-1}_{\theta^\star} \sum_{z_i \in \mathcal{D}} \nabla_\theta \ell(z_i, \wsdmnew{\theta^\star})  \epsilon_i  \Big]^2 + \lambda \| \mathbf{\epsilon} \|_2,
\end{aligned}
\end{footnotesize}
\end{equation}
The first and second terms aim at balancing the TPR and TNR between groups with $\mathbf{\epsilon}$. And the last one is the regularization with weight factor $\lambda\in\mathbb{R}_{+}$, which enforces the weights close to zero as assumed by the influence function~\cite{koh2017understanding}.

\header{Stage Two.}
Next, we train the final model $f_{\theta^{\star}_{\boldsymbol{\epsilon^{\star}}}}$  by reweighting the training samples with $\epsilon^{\star}$.
The weighted loss is,
\begin{equation}
\label{equ:final_goal}
\begin{aligned} 
\wsdmnew{\mathcal{L}}_{\boldsymbol{\epsilon^{\star}}}(\mathcal{D},\theta) = \sum_{i=1}^n (\frac{1}{n} + \epsilon_i^{\star})~l\left(z_i, \theta\right).
\end{aligned}
\end{equation}
We present the detailed  \fairIF~ training algorithm in Appendix~\ref{apd:Algorithm}, Algorithm~\ref{alg:fairif}. Initially, the model is rigorously trained using standard empirical risk until reaching convergence, resulting in the parameter $\theta^{\star}$. Subsequent to this, we calculate the influence function and determine the appropriate weights to achieve equality under metrics TPR and TNR across groups in the validation set. In the final step, we train the model with reweighted data. Note, differently from previous methods~\cite{ren2020notall,teso2021interactive} computing influence function on the fly, in our method, the influence function is computed after the model is converged, which not only saves the computation time of influence function, but also delivers a more accurate estimation. We leave the computation details of the influence function in Appendix~\ref{apd:IF_computation}.

While Equation~\eqref{equ:final_eps_loss} effectively reduces disparity on the validation set, largely due to overparameterization~\cite{zhang2021understanding}, its applicability to the testing phase and potential impact on classification performance remain questions. Subsequent sections, Section~\ref{sec:analysis} and \ref{sec:exp}, address these concerns. Specifically, Section~\ref{sec:analysis} demonstrates bounded disparities between validation and testing fairness metrics, and the task performance guarantee on the test set. Section~\ref{sec:exp}, meanwhile, empirically evaluates \fairIF~ across eight datasets and six models.


%% file: 5_analysis.tex
\vspace{-2mm}
\section{Analysis}
\label{sec:analysis}

This section details \fairIF's characteristics, focusing on minimizing True Positive Rate (TPR) and True Negative Rate (TNR) discrepancies between groups (Section~\ref{sec:disparity_analysis}). This approach addresses three fairness notions: AD, AOE, and AOD.
By exploring the root causes of these discrepancies, we demonstrate that effectively minimizing the disparity in TPR and TNR can alleviate these fairness concerns. The rationale presented solidifies the optimization objective chosen by \fairIF. 
Section~\ref{sec:fair_guarantee} examines how fairness in the validation phase, achieved through optimized sample weights, extends to the testing phase. This is analyzed using TPR and TNR disparities and the Rademacher complexity. In Section~\ref{sec:acc_guarantee}, we showcase the difference in classification accuracy between the original and the retrained models remains minimal when the model converges.

\vspace{-2mm}
\subsection{Mitigating Disparity under Different Notions}
\label{sec:disparity_analysis}
\begin{figure}[t]
    \centering
    \includegraphics[width=0.7\linewidth]{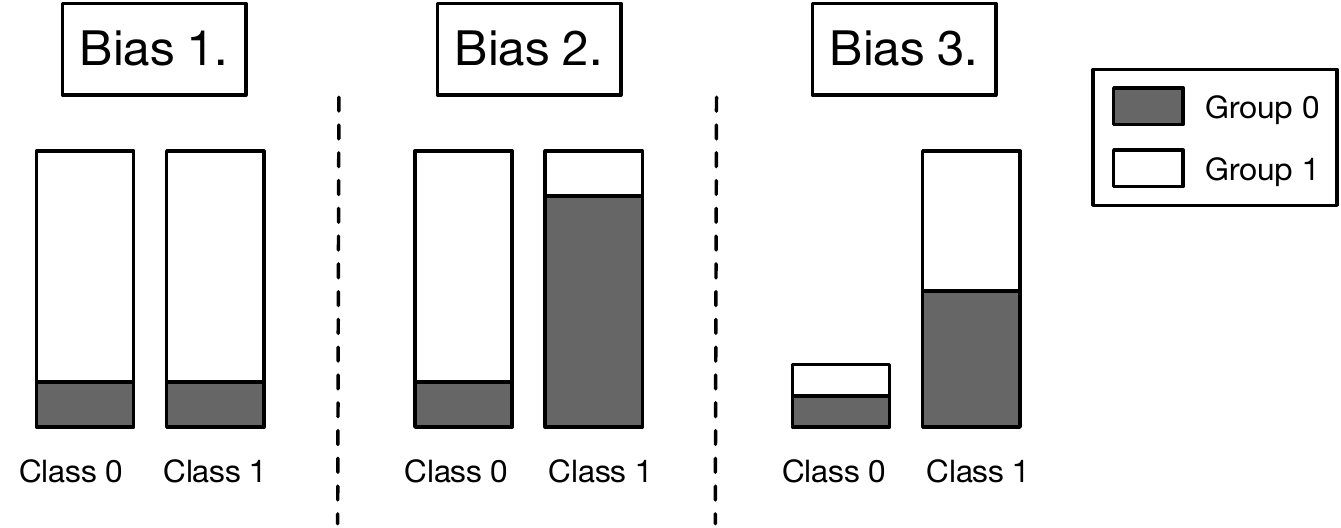}
    \vspace{-10pt}
    \caption{Illustration of three types of bias.}
    \label{fig:bias}
\end{figure}

In this section, we delve into an analytical examination underscoring the significance of balancing True Positive Rates (TPR) and True Negative Rates (TNR) across varied groups. The TPR and TNR equalization is pivotal in alleviating disparities in line with several fairness definitions. Consistent with the findings of prior research~\cite{yu2021fair,reddy2021benchmarking}, the ingrained bias in the models predominantly emerges from the following sources within the training data: 1. Variances in group sizes; 2. Discrepancies in class distribution within individual groups, often termed as the group distribution shift; 3. Inequalities in class sizes. To offer a clearer perspective on these bias forms, refer to Figure~\ref{fig:bias} where we have depicted a visualization of these three distinct bias categories.
With analysis of the relation between the three fairness notions and the TPR and TNR metrics, we derive the following proposition: 
\begin{prop}
\label{them:equ_tpr_tnr}
Under the existence of three types of data bias, if the model prediction satisfies equalized TPR and TNR, i.e., $TPR^0 - TPR^1 = 0$ and $TNR^0 - TNR^1 = 0$, then the three notions of fairness can be achieved, $AD = AOD = EOD = 0$.
\end{prop}

{\it Proof sketch. Given the definitions of Average Odds Difference (AOD) and Equality of Opportunity Difference (EOD) detailed in Section~\ref{sec:notation}, it becomes straightforward to deduce that both AOD and EOD would be zero if the TPR and TNR are harmonized between two distinct groups. Let's delve into the Accuracy Difference (AD). It can be expressed as:
\begin{equation*}
\begin{small}
\vspace{-2mm}
\begin{aligned}
{ AD = |\alpha TPR^{(0)}-\beta TPR^{(1)}+(1-\alpha)TNR^{(0)}-(1-\beta) TNR^{(1)} |},
\end{aligned}
\vspace{-1mm}
\end{small}
\end{equation*} 
Here, \(\mathbb{P}(y=1| s=0)\) and \(\mathbb{P}(y=1| s=1)\) are represented by the variables \(\alpha\) and \(\beta\), respectively. Further dissecting the three forms of bias on a case-by-case basis, it becomes evident that AD either mirrors or is constrained by the combination of differences in TPR and TNR across these scenarios. The comprehensive proof of Proposition~\ref{them:equ_tpr_tnr} has been catalogued for reference in Appendix~\ref{apd:proof}.
}

The aforementioned proposition demonstrates that by harmonizing the TPR and TNR across two groups, we can simultaneously achieve three pivotal notions of fairness, namely: Accuracy Equality, Equal Odds, and Equal Opportunity. This rationale supports our decision to employ a metric that combines both TPR and TNR within \fairIF.

\subsection{Fairness Guarantee on Test Data}
\label{sec:fair_guarantee}
{
In the preceding section, we show that the metric combining TPR and TNR serves as an effective measure of fairness, and our proposed method aspires to ensure equality in TPR and TNR across distinct groups. In this section, our focus shifts to substantiating that the fairness performance remains consistent during testing. Before delving into the fairness assurances offered by \fairIF, it's essential to familiarize ourselves with the concept of Rademacher complexity\cite{bartlett2002localized}, which measures the learnability of function classes.
The Rademacher complexity for a function class is defined as below:
\begin{definition}
Given a space \(Z\), and a set of i.i.d. examples S = \(\{z_1, z_2,...,z_m\} \subseteq Z\), for a function class \(\mathcal{F}\) where each function \(r: Z \rightarrow \mathbb{R}\), the empirical Rademacher complexity of \(\mathcal{F}\) is given by:
\begin{equation}
    \widehat{\operatorname{Rad}}_{S}(\mathcal{F}) = \mathbb{E}_{\sigma}\left[\sup _{r \in \mathcal{F}}\left(\frac{1}{m}\sum_{i=1}^m\sigma_i r(z_i)\right)\right]
\end{equation}
Here, \(\sigma_1,...,\sigma_m\) are independent random variables uniformly drawn from \(\{-1,1\}\).
\end{definition}
In what follows, we establish the bound for TPR disparity during test. A parallel argument can be made for TNR. As indicated in Section~\ref{sec:pre}, the classifier is denoted by \( h(x) \). However, diverging from prior notation for the sake of derivation convenience in this section, we modify the binary output of the classifier to range in \(\{-1, 1\}\). Its corresponding space, \( \mathcal{H} \), encompasses hypotheses with values in \(\{-1, 1\}\). In order to align with the definition of Rademacher complexity, we represent the TPR disparity between two groups on dataset \( \mathcal{D} \) as \( r_{h}(\mathcal{D}) \). Here, the TPR difference \( r_{h} \) acts as a function dependent on the classifier function \( h \). Consequently, the function class \( \mathcal{F} \) is articulated as:
\begin{small}
\begin{align*}
\mathcal{F}=L(\mathcal{H}) \triangleq \Biggl\{ r_{h}(\mathcal{D}) \rightarrow 
&\left| \frac{1}{|\mathcal{D}^{1, y=1}|} \sum_{(x,y=1)\in \mathcal{D}^1} \mathbbm{1}_{\{h(x) = 1\}} - \right.\\
&\left.\left. \frac{1}{|\mathcal{D}^{0, y=1}|} \sum_{(x,y=1)\in \mathcal{D}^0} \mathbbm{1}_{\{h(x) = 1\}} \right|: h \in \mathcal{H}\right\}.
\end{align*}
\end{small}
where \(\mathcal{D}^{1, y=1}\) represent the positive samples from group 1 within \(\mathcal{D}\), while \(\mathcal{D}^{0, y=1}\) signifies the equivalent for group 0. From this definition, it's clear that the range of \( \mathcal{F} \) is bounded within \([0, 1]\). 
}

{
For simplicity in notation, we use \(S\), rather than \(\mathcal{D}^S\), to represent the validation set utilized by our method for sample weight computation. And the size of sample set \(S\) is \(m\). Then, let's define \(R(h) = \mathbbm{E}_{\mathcal{D}^\text{test}}[r_h(\mathcal{D}^\text{test})]\) as the TPR fairness risk of the classifier \(h\) on the test set. Meanwhile, \(\hat{R}_{S}(h) = \frac{1}{m}\sum_{z_i \in S} r_{h}(z_i)\) represents the empirical fairness risk of the classifier \(h\) on the sample set \(S\). Note, samples within \(S\) and \(\mathcal{D}^\text{test}\) are i.i.d. and originate from the same data distribution.
Based on the property of Rademacher complexity \cite{shalev2014understanding},
for any $\delta>0$, with probability at least $1-\delta$ over $S$:
\begin{equation}
\label{equ:defrad}
\forall h \in \mathcal{H}: \quad R(h) \leq \hat{R}_{S}(h)+2* \underline{\widehat{Rad}_{S}(L(\mathcal{H}))}+3 \sqrt{\frac{\log (2 / \delta)}{2 m}}.
\end{equation}
Before we discuss the details of each terms of the previous inequality, let's denote $S^{+}_1$ and $S^{+}_0$ as the positive samples in group 1 and group 0 in the sample set $S$ respectively.
Then for the second term on the right-hand side (RHS) of Equation (\ref{equ:defrad}),  we can express it in terms of  $\widehat{\operatorname{Rad}}_{S}(\mathcal{H})$:
\begin{footnotesize}
{\allowdisplaybreaks
\begin{align*}
\label{equ:express}
&\widehat{\operatorname{Rad}}_{S}(L(\mathcal{H}))=\widehat{\operatorname{Rad}}_{S}(\mathcal{F}) \\
=&\mathbb{E}_{\sigma}\left[
\sup _{h \in \mathcal{H}} \frac{1}{|S^{+}_1|} \sum_{x_i \in S^{+}_1} \sigma_{i} \frac{1+ h\left(x_{i}\right)}{2}
-
\frac{1}{|S^{+}_0|} \sum_{x_i \in S^{+}_0} \sigma_{i} \frac{1+ h\left(x_{i}\right)}{2}
\right] \\
\leq& \mathbb{E}_{\sigma}\left[\sup _{h \in \mathcal{H}}
\frac{1}{|S^{+}_1|} \sum_{x_i \in S^{+}_1} \sigma_{i} \frac{1+ h\left(x_{i}\right)}{2}
\right] + \mathbb{E}_{\sigma}\left[\sup _{h \in \mathcal{H}}
\frac{1}{|S^{+}_0|} \sum_{x_i \in S^{+}_0} \sigma_{i} \frac{1+ h\left(x_{i}\right)}{2}
\right] \\
=& \mathbb{E}_{\sigma}\left[\frac{1}{2 |S^{+}_1|} \sum_{i=1}^{|S^{+}_1|} \sigma_{i}+\frac{1}{2} \sup _{h \in \mathcal{H}} \frac{1}{|S^{+}_1|} \sum_{x_i \in S^{+}_1} \sigma_{i}  h\left(x_{i}\right)\right] \\
&+\mathbb{E}_{\sigma}\left[\frac{1}{2 |S^{+}_0|} \sum_{i=1}^{|S^{+}_0|} \sigma_{i}+\frac{1}{2} \sup _{h \in \mathcal{H}} \frac{1}{|S^{+}_0|} \sum_{x_i \in S^{+}_0} \sigma_{i}  h\left(x_{i}\right)\right]
\\
=&\frac{1}{2} \mathbb{E}_{\sigma}\left[\sup _{h \in \mathcal{H}} \frac{1}{|S^{+}_1|} \sum_{x_i \in S^{+}_1} \sigma_{i}  h\left(x_{i}\right)\right]
+ \frac{1}{2} \mathbb{E}_{\sigma}\left[\sup _{h \in \mathcal{H}} \frac{1}{|S^{+}_0|} \sum_{x_i \in S^{+}_0} \sigma_{i}  h\left(x_{i}\right)\right] 
=\underline{\widehat{\operatorname{Rad}}_{S}(\mathcal{H})}
\end{align*}
}
\end{footnotesize}
Our theoretical analysis elucidates key insights from Equation (\ref{equ:defrad}). Specifically, the second term on the right-hand side primarily reflects the Rademacher complexity of the model. This complexity is intricately connected to the family of neural networks to which the model belongs, and is constrained by moderate assumptions~\cite{wei2018margin}.
Concurrently, due to the power of over-parameterization~\cite{zhang2021understanding}, the first term, \(\hat{R}_{S}(h)\), approaches zero (the discrepancy on the validation set can be optimized to nearly vanish).
The third term's magnitude correlates with the validation set size \(m\), and as this size expands, our bound tightens. It's noteworthy that for current deep learning datasets, a validation set, even if relatively small compared to the training set, is often ample. Empirical validations of this claim are further explored in Appendix~\ref{apd:valid_size_exp}.
}

\subsection{Accuracy Guarantee on Test Data}
\label{sec:acc_guarantee}
In this subsection, our primary objective is to understand the model's task performance on the test dataset. To achieve this, we represent the test loss, using $\theta^{\star}_{{\epsilon}}$ for each sample, leveraging the first-order Taylor approximation:
\begin{equation}
\label{equ:acc_test}
  \ell\left(z_{\text {test}}, \wsdmnew{\theta^{\star}_{{\epsilon}}}\right) = \ell\left(z_{\text {test}}, \wsdmnew{\theta^{\star}}\right) + \mathcal{I}_{\text {loss }}\left(z_{\text {test}}, \mathcal{D} \right) \epsilon + \mathcal{O}\left(\|\epsilon\|^{2}\right).
\end{equation}
For many benchmark datasets, $\|\epsilon\|^{2} \leq \frac{c}{n^{2}}$ holds~\cite{yang2022dataset}, where $c$ represents the data quantity for which $\epsilon_i \neq 0$. 
Following the Influence Function methodology in \cite{koh2017understanding}, we omit the term $\mathcal{O}\left(\|\epsilon\|^{2}\right)$ in Equation (\ref{equ:acc_test}). Expanding this further to encompass the test loss over the test set $\mathcal{D}^\text{test}$, we derive:

\begin{equation}
\label{equ:proof2}
\begin{small}
\begin{aligned}
\mathcal{L}(\mathcal{D}^\text{test}&, {\theta}^{\star}_{{\epsilon}}) - \mathcal{L}\left(\mathcal{D}^\text{test},{\theta^{\star}}\right) \approx \mathbb{E}_{z_{\text {test }} \in \mathcal{D}^\text{test}}\left[\mathcal{I}_{\text {loss}}\left(z_{\text {test }}, {\mathcal{D}}\right)\right] \epsilon \\
&= \left[-\nabla_{\theta} \wsdmnew{\mathcal{L}}\left(\mathcal{D}^\text{test}, {\theta^{\star}}\right)^{\top}\right]\left[\sum_{{z}_{i} \in {\mathcal{D}}} H_{{\theta^{\star}}}^{-1} \nabla_{\theta} \wsdmnew{\ell}\left({z}_{i}, {\theta^{\star}}\right)\right] \epsilon \\
& \leq \left\|\nabla_{\theta} \mathcal{L}(\mathcal{D}^\text{test},{\theta^{\star}})\right\|_{2}\left\|\sum_{{z}_{i} \in {\mathcal{D}}} \mathcal{I}_{\text {param }}\left({z}_{i}\right)\right\|_{2} \left\| \epsilon \right\|_2\\
& \leq \left\|\nabla_{\theta} \mathcal{L}(\mathcal{D}^\text{test},{\theta^{\star}})\right\|_{2}\ \cdot \gamma \cdot \left\| \epsilon \right\|_2
\end{aligned}
\end{small}
\end{equation}

The concluding inequality assumes a positive real number $\gamma$ exists such that $\left\|\sum_{{z}_{i} \in {\mathcal{D}}} \mathcal{I}_{\text {param }}\left({z}_{i}\right)\right\|_{2} \leq \gamma$. Even though this assumption aligns with earlier works~\cite{yang2022dataset}, it's crucial to note that $\left\|\sum_{{z}_{i} \in {\mathcal{D}}} \mathcal{I}_{\text {param }}\left({z}_{i}\right)\right\|_{2} \leq \sum_{{z}_{i} \in {\mathcal{D}}} \left\|\mathcal{I}_{\text {param }}\left({z}_{i}\right)\right\|_{2}$. This term signifies the influence of each training sample on the test loss.
Given that empirically introducing or excluding individual one training sample only marginally impacts the test loss, the assumption of $\gamma$ is validated. Moreover, with the objective pushing $\left\| \epsilon \right\|_2$ towards zero, variations in test performance between our fair model and the original unweighted model are constrained.

%% file: 6_experiments.tex
\section {Experiments} 
\label{sec:exp}

\begin{table*}[!t]
\centering
\caption{(MLP) Comparison of \fairIF\ with baselines on CI-MNIST with different types of bias. The Original row shows the performances of the model trained with ERM. For AD, AOD and EOD, smaller values indicate better fairness performance. Bold font is used to highlight the best result and the underscores are for the second-best result.}
\vspace{-0.4cm}
\scalebox{0.88}{
\begin{tabular}{l | cccc | cccc | cccc}
\toprule
& \multicolumn{4}{c|}{\bf Group Size Discrepancy} & \multicolumn{4}{c}{\bf Group Distribution Shift} & \multicolumn{4}{|c}{\bf Class Size Discrepancy}\\
\hline
Models &{\bf Acc}(\%)&{\bf AD}(\%) &{\bf AOD} &{\bf EOD} &{\bf Acc}(\%)&{\bf AD}(\%) &{\bf AOD} &{\bf EOD} &{\bf Acc}(\%)&{\bf AD}(\%) &{\bf AOD} &{\bf EOD} \\
\hline
Original & \underline{97.55} & 1.104 & {0.013} & \underline{0.011} & \underline{98.19} & 0.256 & 0.065 & 0.056 & \underline{97.27} & 0.818 & 0.010 & 0.012  \\
\hline
CFair & 97.09 & 1.745 & 0.018 & 0.026 & 97.49 & \textbf{0.060} & \underline{0.041} & {0.034} & 96.09 & 0.798 & 0.009 & \underline{0.011}  \\
DOMIND & 92.96 & 1.162 & 0.020 & 0.031 & 97.23 & 0.696 & 0.074 & 0.047 & 96.64 & 0.798 & 0.027 & 0.051  \\
ARL & 96.95 & 1.424 & 0.014 & 0.017 & 97.62 & 0.374 & 0.085 & 0.070 & 96.54 & 0.335 & \underline{0.008} & 0.015  \\
FairSMOTE & 96.18 & {1.031} & {0.014} & 0.016 & 97.36 & {0.292} & 0.054 & {0.038} & 96.25 & \underline{0.310} & \textbf{0.006} & \textbf{0.010}  \\
Influence &97.40 & \textbf{0.708} & \textbf{0.007} &0.010 &97.11 &0.760 & \textbf{0.037} & \textbf{0.009}& 97.25 & 0.443& 0.010 & 0.019\\
\hline
\fairIF & \textbf{98.01} & \underline{1.012} & \underline{0.011} & \textbf{0.006} & \textbf{98.49} & \underline{0.223} & {0.053} & \underline{0.031} & \textbf{97.71} & \textbf{0.108} & \underline{0.008} & \underline{0.011}\\  
\bottomrule
\end{tabular}
}
\label{tab:cl_mnist_mlp}
\end{table*}

\begin{table*}[t]
\centering
\caption{(CNN) Comparison of \fairIF\ with baselines on CI-MNIST with different types of bias. Bold font highlights the best result and the underscores are for the second-best result.}
\vspace{-0.4cm}
\scalebox{0.88}{
\begin{tabular}{l | cccc | cccc | cccc}
\toprule
& \multicolumn{4}{c|}{\bf Group Size Discrepancy} & \multicolumn{4}{c}{\bf Group Distribution Shift} & \multicolumn{4}{|c}{\bf Class Size Discrepancy}\\
\hline
Models &{\bf Acc}(\%)&{\bf AD}(\%) &{\bf AOD} &{\bf EOD} &{\bf Acc}(\%)&{\bf AD}(\%) &{\bf AOD} &{\bf EOD} &{\bf Acc}(\%)&{\bf AD}(\%) &{\bf AOD} &{\bf EOD} \\
Original & 98.46 & 1.090 & 0.011 & 0.011 & 98.81 & 0.577 & 0.043 & 0.030 & 98.65 & 0.488 & 0.006 & 0.008  \\
\hline
CFair & 98.62 & \textbf{0.233} & \textbf{0.002} & \textbf{0.001} & 97.17 & 2.755 & \underline{0.007} & 0.008 & 98.37 & \textbf{0.095} & \textbf{0.002} & \textbf{0.003}    \\
DOMIND & \textbf{98.97} & 0.779 & 0.008 & 0.015 & \underline{98.88} & 0.248 & \underline{0.007} & \underline{0.004} & \textbf{98.92} & 0.437 & 0.006 & 0.009  \\
ARL & 98.35 & 0.999 & 0.010 & 0.012 & 98.69 & 0.599 & 0.049 & 0.028 & 98.43 & 0.482 & \underline{0.004} & \textbf{0.003}  \\
FairSMOTE & 97.94 & {0.541} & 0.007 & 0.008 & 98.58 & {0.241} & 0.009 & 0.006 & 97.92 & 0.383 & 0.005 & \underline{0.004}  \\
Influence & 98.13 & 0.468 & \underline{0.006} & \underline{0.003} & 98.42 &  \underline{0.235} & 0.049 & 0.037 & 97.99 & 0.475 & 0.005 & \textbf{0.003}\\
\hline
\fairIF & \underline{98.69} & \underline{0.405} & \underline{0.006} & {0.007} & \textbf{98.94} & \textbf{0.234} & \textbf{0.004} & \textbf{0.003} & \underline{98.80} & \underline{0.317} & 0.005 & \underline{0.004}\\
\bottomrule
\end{tabular}
}
\label{tab:cl_mnist_cnn}
\end{table*}

\begin{table*}[t]
\centering
\caption{(LeNet) Comparison of \fairIF\ with baselines on CI-MNIST with different types of bias. Bold font highlights the best result and the underscores are for the second-best result.} 
\vspace{-0.4cm}
\scalebox{0.88}{
\begin{tabular}{l | cccc | cccc | cccc}
\toprule
& \multicolumn{4}{c|}{\bf Group Size Discrepancy} & \multicolumn{4}{c}{\bf Group Distribution Shift} & \multicolumn{4}{|c}{\bf Class Size Discrepancy}\\
\hline
Models &{\bf Acc}(\%)&{\bf AD}(\%) &{\bf AOD} &{\bf EOD} &{\bf Acc}(\%)&{\bf AD}(\%) &{\bf AOD} &{\bf EOD} &{\bf Acc}(\%)&{\bf AD}(\%) &{\bf AOD} &{\bf EOD} \\
\hline
Original & 99.11 & 0.464 & 0.005 & 0.006 & 99.17 & 0.153 & 0.026 & 0.018 & 98.81 & 0.152 & \textbf{0.001} & \textbf{0.001}  \\
\hline
CFair & 99.07 & \textbf{0.204} & 0.005 & 0.007 & 99.16 & 0.118 & \underline{0.020} & \textbf{0.012} & 98.65 & 0.127 & 0.003 & 0.004  \\
DOMIND & \underline{99.12} & 0.318 & \underline{0.004} & 0.007 & \textbf{99.54} & \underline{0.058} & 0.021 & 0.014 & \underline{99.12} & 0.177 & 0.004 & 0.006  \\
ARL & 99.10 & 0.278 & \textbf{0.003} & {0.004} & 98.86 & 0.172 & 0.034 & 0.028 & 98.69 & 0.266 & 0.003 & 0.004  \\
FairSMOTE & 98.96 & \underline{0.226} & \underline{0.004} & 0.006 & 99.03 & 0.091 & 0.022 & 0.021 & 98.74 & {0.118} & 0.003 & 0.004  \\
Influence & 98.91 & 0.287 & \textbf{0.003} & \textbf{0.001} & 98.67 & 0.168 & 0.042 & 0.030 &99.05 & \textbf{0.064} & 0.004 &0.004\\
\hline
\fairIF & \textbf{99.13} & 0.375 & \underline{0.004} & \underline{0.003} & \underline{99.42} & \textbf{0.042} & \textbf{0.018} & \underline{0.013} & \textbf{99.25} & \underline{0.088} & \underline{0.002} & \underline{0.003} \\
\bottomrule
\end{tabular}
}
\label{tab:cl_mnist_lenet}
\end{table*}
In our experiments, we initially underscore the effectiveness of \fairIF\ in mitigating fairness concerns arising from three distinct bias types. This is illustrated using the synthetic dataset, CI-MNIST (Section~\ref{sec:syn_exp}). Subsequently, we compare \fairIF\ against five baseline approaches, effectively illustrating the method's performance and scalability on real-world datasets (Section~\ref{sec:real_exp}). We further showcase the empirical results obtained when integrating \fairIF\ with pre-existing models (Section~\ref{sec:pretrain_exp}). Finally, we examine the data instances that undergo reweighting in the training set to achieve fairness, assessing their alignment with human preference (Section~\ref{sec:weight_study}).
We leave the empirically investigation of the size of validation set to \fairIF's operation in Appendix~\ref{apd:valid_size_exp}.

Regarding datasets employed, our experiments leverage three variations of the synthetic image dataset \textbf{CI-MNIST}\cite{reddy2021benchmarking}. Additionally, three real-world tabular datasets: \textbf{Adult}\cite{asuncion2007uci}, \textbf{German}\cite{asuncion2007uci}, and \textbf{COMPAS}\cite{dieterich2016compas}, along with two genuine image datasets, \textbf{CelebA} \cite{liu2015deep} and \textbf{FairFace}\cite{karkkainen2019fairface}, are integrated into our study.
For the comparative baselines, we utilize the following approaches: \textbf{CFair}\cite{zhao2019conditional}, \textbf{DOMIND}\cite{wang2020towards}, \textbf{ARL}\cite{lahoti2020fairness}, \textbf{FairSMOTE}\cite{chakraborty2021bias}, and \textbf{Influence} \cite{li2022achieving}. We leave the details about datasets, baselines,  scalable implementation of influence function, and configuration details in appendix~\ref{apd:IF_computation}, ~\ref{apd:baseline_detail}, ~\ref{apd:dataset_detail}, and~\ref{apd:implementataion}.

\vspace{-2mm}
\subsection{Synthetic Experiments}
\label{sec:syn_exp}
The synthetic dataset CI-MNIST~\cite{reddy2021benchmarking} is a variant of the MNIST dataset. In this dataset, each input image $x$ has a label $y \in \{0,1\}$ indicating odd or even respectively, and the background color, blue or red, is the sensitive attribute. CI-MNIST offers manual control over different types of bias in the dataset, such as the number of samples in each group and class, and the group distribution over the class. To examine the effectiveness of \fairIF~under different types of bias, we independently introduce the three major types of bias analyzed in Section~\ref{sec:analysis} into the dataset. 
We compare \fairIF~with the five other baselines on three CI-MNIST variants based on three models: a multilayer perception (MLP), a convolutional network (CNN) and a LeNet\cite{lecun1998gradient}. 
The experimental results are shown in Table~\ref{tab:cl_mnist_mlp}, Table \ref{tab:cl_mnist_cnn} and Table \ref{tab:cl_mnist_lenet} respectively. We report the results of the epochs with the best task accuracy performance. We provide the visualization of three different types of data bias in Figure~\ref{fig:bias}.\\
\textbf{Bias 1: Group Size Discrepancy.} We set 15\% of images in the two classes as red and the remaining as blue. The distributions over classes of each group are the same, and the number of images in each class is also the same. In this setting, the group of red background is under-representative.\\
\textbf{Bias 2:  Group Distribution Shift.} We keep the amount of data within each class and each group to be the same, but set 85\% of images in class 0 with blue background and 15\% of images in class 1 with blue background. In this case, the group distributions over the classes are different. \\
\textbf{Bias 3: Class Size Discrepancy.} We set group distributions and the total amount of data within each group to be the same. And the amount of data in class 1 is 25\% of class 0.

\begin{table*}[t]
\centering
\caption{Comparison of \fairIF\ with baselines on three real-world tabular datasets. Bold font is used for the best values.} 
\vspace{-0.4cm}
\scalebox{0.88}{
\begin{tabular}{l | cccc | cccc |cccc}
  \bottomrule
  & \multicolumn{4}{c|}{\bf Adult}   & \multicolumn{4}{c|}{\bf German}  & \multicolumn{4}{c}{\bf COMPAS}\\
\hline
Methods
&{\bf Acc}(\%)&{\bf AD}(\%) &{\bf AOD} &{\bf EOD}
&{\bf Acc}(\%)&{\bf AD}(\%) &{\bf AOD} &{\bf EOD}
&{\bf Acc}(\%)&{\bf AD}(\%) &{\bf AOD} &{\bf EOD} \\
 \hline
Original & 69.76 & 17.40 & 0.302 & 0.303 & {\bf 75.50} & 15.45 & 0.083 & 0.041 & {\bf 68.09} & 2.09 & 0.183 & 0.220\\
\hline
CFair  & {\bf 70.78} & 16.97 & 0.312 & 0.345 & 69.50 & 11.01 & 0.018 & {\bf 0.020} & 66.29 & 0.29 & 0.399 & 0.455  \\
DOMIND  & 68.51 & 15.92 & 0.222 & {\bf 0.237} & 69.00 & 10.26 & 0.072 & 0.103 & 67.89 &{0.36} & 0.174 &0.240 \\
ARL & 69.49 & 20.53 & 0.373 &0.388 & 74.50 & 12.16 & 0.027 & 0.036 &  64.30 & 0.65&0.197 &0.269  \\
FairSMOTE  & 68.43 & 16.67 & 0.268 & 0.271 & 64.12 & 9.92 & 0.053 & 0.082 &66.14 & 1.42& 0.198 & 0.242  \\
Influence & 68.79 & 15.85 & 0.269&0.268 & 72.01 & 13.25 & 0.056 & 0.107 &67.32 & 1.01 & 0.164 & \textbf{0.163}\\
\hline
\fairIF & 68.97 & {\bf 15.34} & {\bf 0.170} & 0.265 & {74.68} & {\bf 8.68} & {\bf 0.017} & 0.024 & 67.53 & {\bf 0.28} & {\bf 0.156} &  0.218 \\
\toprule
\end{tabular}
}
\vspace{-2mm}
\label{tab:real_tab_dataset}
\end{table*}

\begin{table}[t]
\centering
\small
\caption{Comparison of \fairIF\ with baselines on two real-world image datasets. Bold font is used for the best values.}
\vspace{-0.4cm}
        \scalebox{0.85}{
        \setlength{\tabcolsep}{3pt}{
\begin{tabular}{l | cccc |cccc}
  \bottomrule
& \multicolumn{4}{c|}{\bf FairFace} & \multicolumn{4}{c}{\bf CelebA}\\
\hline
Methods &{\bf Acc}(\%)&{\bf AD}(\%) &{\bf AOD} &{\bf EOD}&{\bf Acc}(\%)&{\bf AD}(\%) &{\bf AOD} &{\bf EOD} \\
 \hline
Original & 86.68 & 7.23 &  0.0721& 0.0610  & {\bf 95.41} & 4.54 & 0.2485 & 0.4975\\
\hline
CFair & {\bf 86.72} & 5.12& 0.0509 &  0.0358& 95.11 &  4.23& 0.1954 & 0.3520 \\
DOMIND & 86.69 & 4.86 & 0.0496 & 0.0172 & 94.88 & 4.18  &0.1864  & 0.3475 \\
ARL & 86.59 &5.59 & 0.0078 & 0.0125  & 95.08 & 4.31 & 0.2185 &  0.4108  \\
FairSMOTE & 86.42& 5.86 & 0.0488 & 0.0267 & 95.26 & 4.42 & 0.2256 & 0.3872 \\
\hline
\fairIF &  {86.63} & {\bf 4.14} & {\bf 0.0048} & {\bf 0.0105} & {95.37} & {\bf 3.81} & {\bf 0.1220} & {\bf 0.3145} \\
\toprule
\end{tabular} }}
\vspace{-2mm}
\label{tab:real_dataset}
\end{table}

\begin{table}[h!]
\centering
\small
\caption{Performance of \fairIF~with Pretrained Models. The relative change caused by using \fairIF~is presented in percentage. \underline{Smaller} values for AD, AOD, and EOD indicate larger discrepancy mitigation.}
\vspace{-0.4cm}
        \scalebox{0.85}{
        \setlength{\tabcolsep}{3pt}{
\begin{tabular}{l | llll | llll}
  \bottomrule
& \multicolumn{4}{c|}{\bf FairFace} & \multicolumn{4}{c}{\bf CelebA}\\
\hline
Models &{\bf Acc}&{\bf AD} &{\bf AOD} &{\bf EOD} &{\bf Acc}&{\bf AD} &{\bf AOD} &{\bf EOD} \\
\hline
+ResNet-18 & +0.40\%  & -35.81\% & -48.81\% & -66.87\% & +0.07\%  & -23.73\% & -37.08\% & -23.51\% \\
+ResNet-34 & -0.17\% & -21.05\% & -77.96\% & -77.76\% & -0.21\% & -9.66\%  & -25.78\%  & -7.80\% \\
+ResNet-50 & -0.01\% & -9.46\%  & -37.21\% & -49.90\% & -0.34\% & -29.02\% & -28.22\% & -9.76\% \\
\toprule
\end{tabular} }}
\label{tab:pretrain}
\end{table}

As shown in Table~\ref{tab:cl_mnist_mlp}, \ref{tab:cl_mnist_cnn} and \ref{tab:cl_mnist_lenet}, compared with the original model trained by ERM, \fairIF~can achieve lower discrepancies under three different notions of fairness, which demonstrates that \fairIF~can mitigate the fairness issue caused by the three different types of bias. In the meantime, we notice that \fairIF~mostly achieves higher accuracy than the original model under different biases, indicating the spurious correlation problem is alleviated by our sample reweighting mechanism. Also, we observe that the Influence baseline often achieves a high fairness level but at an unsatisfactory sacrifice of task performance, although we have conducted a hyperparameter grid search for the best fairness-utility trade-off. Further, comparing \fairIF~with all other state-of-the-art methods, we find that \fairIF~mostly has the top two performance on all the metrics regardless of the model and the dataset.
This shows \fairIF\ can achieve
better fairness-utility trade-offs, even compared with methods directly using all the group information (e.g., FairSMOTE and DOMIND) and additional adversarial models (e.g., CFair and ARL).

\vspace{-2mm}
\subsection{Real-world Experiments}
\label{sec:real_exp}
In this study, we evaluate \fairIF\ in comparison to five baseline models, utilizing three distinct tabular datasets: Adult \cite{asuncion2007uci}, German \cite{asuncion2007uci}, and COMPAS \cite{dieterich2016compas}. To mitigate the risk of overfitting on these datasets, we follow the previous works~\cite{li2022achieving, wang2022understanding} and adopt the logistic regression for testing.
Results can be found in Table \ref{tab:real_tab_dataset}. 
Furthermore, 
to demonstrate the effectiveness and scalability of our proposed method with large datasets and advanced models, we further tests on image datasets CelebA~\cite{liu2015deep} and FairFace~\cite{karkkainen2019fairface} were conducted. For fairness, all baseline models were adapted using ResNet-18 as a feature extractor. It's worth noting that results for the Influence baseline on image datasets were omitted due to prohibitive computational demands. These findings are presented in Table~\ref{tab:real_dataset}. Hyperparameters for all methodologies were tuned based on validation set performance.

As shown in Table~\ref{tab:real_tab_dataset} and \ref{tab:real_dataset}, we find that just a small amount of group information on the validation set can allow \fairIF~to achieve low performance discrepancies between demographic groups. In the meantime, the accuracy performance of \fairIF\ is very close to the original method, which supports our theoretical guarantees. Compared with the Influence baseline, \fairIF\ often achieves a similar or better fairness level while maintaining better task performance on the three tabular datasets. And \fairIF\ is compatible with deep neural networks on large-scale image datasets while the Influence baseline is not.
Furthermore, even comparing with methods that directly use group annotation on the training set, e.g. CFair, DOMIND and FairSMOTE, \fairIF~can still most often outperform them in both accuracy and fairness metrics regardless of the backbone model, which proves its efficacy and scalability.


\vspace{-2mm}
\subsection{Debiasing Pretrained Models}
\label{sec:pretrain_exp}
As observed in previous research~\cite{wang2019racial}, the pretrained models deliver discriminative results across different demographic groups, which might be caused by the pretraining procedure and the pretraining dataset collected from social networks, international online newspapers, and web searches. The proposed method \fairIF~yields fair models through changing the sample weights, which makes \fairIF~ a promising approach to remove the discrimination encoded in the pretrained parameters. Note, for methods requiring a modification of the network structure, the power of pretraining usually cannot be fully utilized.
In this section, we aim to answer the question of how \fairIF~mitigates the fairness issue of the pretrained models. We evaluate \fairIF~with three commonly used pretrained models, ResNet-18, ResNet-34, and ResNet-50~\cite{he2016deep}.
We finetune and evaluate these pretrained models on FairFace and CelebA and compare them with their counterparts trained with \fairIF. The relative changes caused by using \fairIF~are presented in percentage in Table~~\ref{tab:pretrain}. With three different pretrained models on the two datasets, \fairIF~consistently mitigates the discrepancies of three different notions of fairness without hurting the accuracy performance.

\vspace{-2mm}
\subsection{Sample Weight Study}
\label{sec:weight_study}
\begin{figure}[h]
\centering
\vspace{-3mm}
\includegraphics[width=0.89\linewidth]{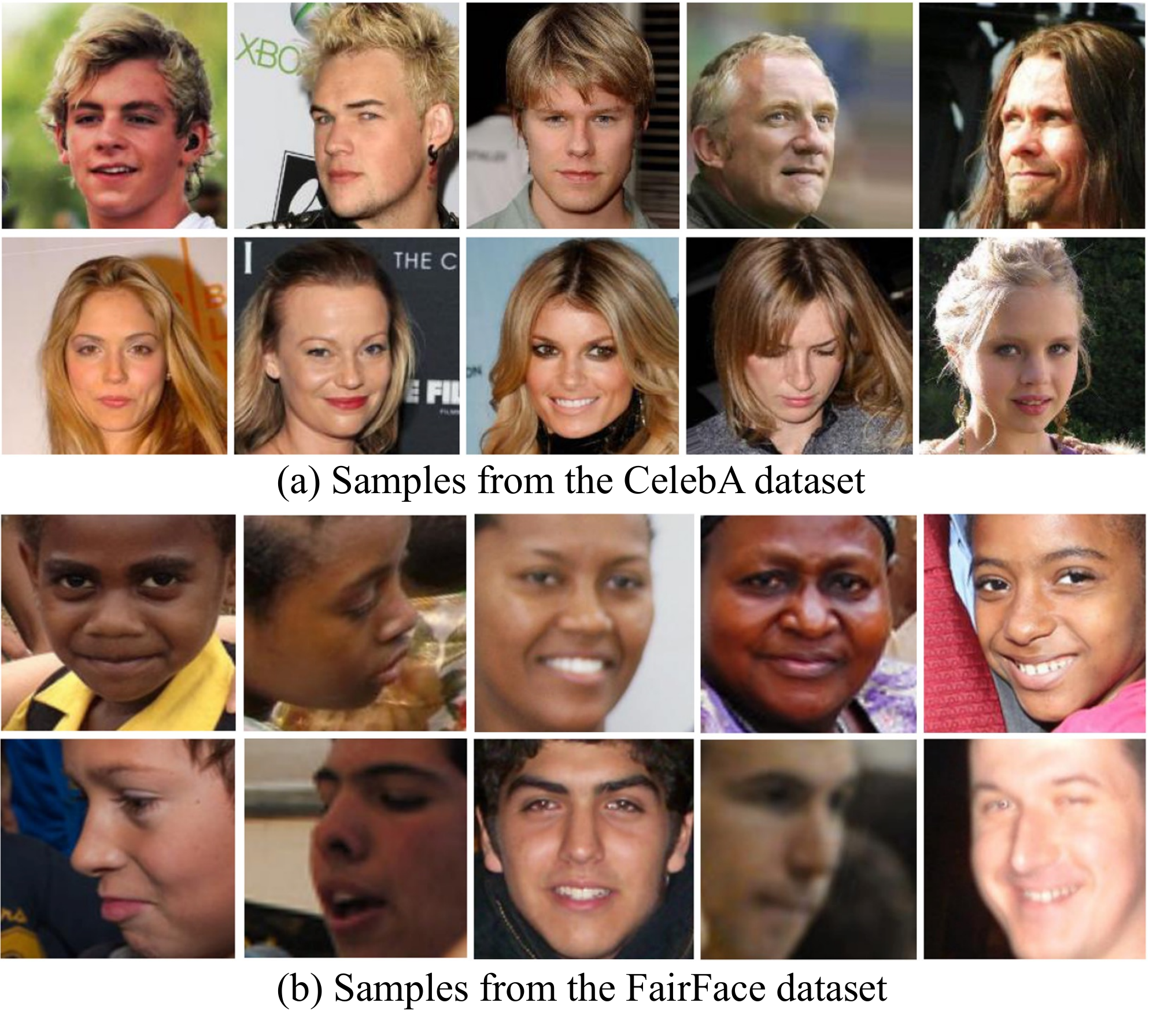}
\vspace{-5mm}
\caption{{Examples of the reweighting done by \fairIF.} In each figure, mostly up-weighted samples are presented in the first row, and mostly down-weighted are in the second.}
\label{fig:weight_study}
\end{figure}

We now probe into how \fairIF~achieves low performance discrepancies across different groups with the same or higher accuracy. In order to perform this analysis, we use the group annotations on the training data to closely examine what examples are up-weighted or down-weighted in the training set. Note, we don't use the group information in the training stage. We show the samples mostly upweights and downweights by \fairIF~in Figure~\ref{fig:weight_study}.
For the CelebA dataset, we observe that \fairIF~tends to give more weight to examples of men with blond hair and white hair, while it decreases the weight of examples of females with blond hair. The result is as expected, the group (Male, Blond Hair) is the minority and the group (Female, Blond Hair) is the majority. 
For the FairFace dataset, \fairIF~ tends to upweight the samples of people with dark skin and downweight the samples of people with white skin. This is aligned with our expectation because the accuracy for group $s=1$ (white) is higher than the group $s=0$ (black). By upweighting the minority group and downweighting the majority group, the  performance tends to balance. Note that \fairIF\ computes different weights for different samples, and it is thus smarter than simply equalizing the weights of the samples from different groups in different classes in other reweighting methods such as FairSMOTE.

%% file: 7_conclusion.tex
\vspace{-5mm}
\section{Conclusion}
\label{sec:con}
Empirical loss minimization in training machine learning models can unintentionally amplify inherent discrimination and societal biases. Recognizing this, we presented \fairIF, a novel two-stage training framework. Distinct from methods that rely heavily on sensitive training data or demand major model alterations, \fairIF\ re-trains on a weighted dataset. These weights, derived using the influence function, ensure uniform model performance across diverse groups. The unique selling point of \fairIF\ is its adaptability: it integrates seamlessly with models using stochastic gradient descent without altering the training algorithm, requiring only group annotations from a small validation set.
Theoretically, we showcased that the performance delta between the reweighted-data model and the original optimal one remains finite, and fairness discrepancies during testing across groups are limited. By addressing these discrepancies, \fairIF\ adeptly tackles disparities stemming from group and class size variations, as well as distribution shifts. 
Empirical assessments on synthetic datasets underscore \fairIF's capability in producing models that better balance fairness and utility. Tests on real-world datasets further vouch for its efficiency and scalability. Additionally, our exploration with pretrained models demonstrates \fairIF's prowess in leveraging their strengths for subsequent tasks, all the while rectifying fairness issues from their training stages.

\vspace{-1.5mm}
\begin{acks}
    This work is supported by National Science Foundation under Award No. IIS-1947203, IIS-2117902, IIS-2137468, and IIS-2002540. The views and conclusions are those of the authors and should not be interpreted as representing the official policies of the funding agencies or the government.
\end{acks}

%% file: 8_appendix.tex
\appendix
\section{APPENDIX: Proof of Proposition}
\label{apd:proof}
{\it Proof.} 
For Equal Odds and Equal Opportunity, recall that they are defined as:
\begin{equation*}
\begin{small}
\begin{aligned}
AOD &=\frac{1}{2}\big[|\text{TPR}^{(1)} - \text{TPR}^{(0)}| + |\text{TNR}^{(1)} - \text{TNR}^{(0)}| \big].\\
EOD &= |\text{TPR}^{(1)} - \text{TPR}^{(0)}|.
\end{aligned}
\end{small}
\end{equation*}
When TPR and TNR are equalized between the two groups, AOD and EOD are both 0, and thus Equal Odds and Equal Opportunity are achieved.

For Accuracy Equality, denoting $\mathbb{P}(y=1| s=0)$ and $\mathbb{P}(y=1| s=1)$ as $\alpha$ and $\beta$ respectively, we can rewrite $AD$ into:
\begin{equation*}
\begin{small}
\begin{aligned}
&\Scale[0.91]{AD = |\alpha TPR^{(0)}-\beta TPR^{(1)}+(1-\alpha)TNR^{(0)}-(1-\beta) TNR^{(1)} |}.
\end{aligned}
\end{small}
\end{equation*}
Next we discuss how equalized TPR and TNR helps achieve Accuracy Equality with the presence of the three types of bias.

\noindent\textit{Bias 1: Group Size Discrepancy.} When the two groups have different group sizes but the same class distribution and the same class size, we have $\alpha = \beta$, and thus 
\begin{equation*}
\begin{small}
\begin{aligned}
\Scale[0.91]{AD} &\Scale[0.91]{=  |\alpha TPR^{(0)}-\alpha TPR^{(1)}+(1-\alpha)TNR^{(0)}-(1-\alpha) TNR^{(1)}|}\\
&\Scale[0.91]{\leq \alpha|TPR^{(0)}- TPR^{(1)}|+(1-\alpha)|TNR^{(0)}- TNR^{(1)}|}.
\end{aligned}
\end{small}
\end{equation*}
\textit{Bias 2:  Group Distribution Shift.} When the two groups have different class distributions but the same group size and class size, we have $\alpha = 1 - \beta$. And without loss of generality, we have $\alpha \leq 0.5 \leq \beta$ (as shown in the middle figure in Figure~\ref{fig:bias}). Then we can get,
\begin{equation*}
\begin{small}
\begin{aligned}
\Scale[0.91]{AD =}&  \Scale[0.91]{|\alpha TPR^{(0)}-(1-\alpha) TPR^{(1)}+(1-\alpha) TNR^{(0)}- \alpha TNR^{(1)}|}\\
& \Scale[0.91]{\leq  \alpha|TPR^{(0)}- TPR^{(1)}| + \alpha |TNR^{(0)}- TNR^{(1)}|}\\
& \Scale[0.91]{+ (1-2\alpha)|TPR^{(1)}- TNR^{(0)}|}.\\
\end{aligned}
\end{small}
\end{equation*}
The group 0 and group 1 dominate two different classes with same proportion. In the example, Figure~\ref{fig:bias}, 85\% of data in class 0 belongs to group 0; and 85\% of data in class 1 belongs to group 1. We assume the difficulty of fitting two classes are same, then $TPR^{(1)} \approx TNR^{(0)}$. We further have,
\begin{equation*}
\begin{small}
\begin{aligned}
AD \leq & \alpha|TPR^{(0)}- TPR^{(1)}|+\alpha|TNR^{(0)}- TNR^{(1)}|.
\end{aligned}
\end{small}
\end{equation*}
\textit{Bias 3: Class Size Discrepancy.} When the two classes have different sizes but the two groups have the same size and class distributions, we have $\alpha = \beta$, and thus 
\begin{equation*}
\begin{small}
\begin{aligned}
\Scale[0.91]{AD} & \Scale[0.91]{=  |\alpha TPR^{(0)}-\alpha TPR^{(1)}+(1-\alpha)TNR^{(0)}-(1-\alpha) TNR^{(1)}|}\\
&\Scale[0.91]{\leq \alpha|TPR^{(0)}- TPR^{(1)}|+(1-\alpha)|TNR^{(0)}- TNR^{(1)}|}.
\end{aligned}
\end{small}
\end{equation*}

\section{APPENDIX: fairIF Algorithm}
\label{apd:Algorithm}
The model initially undergoes training using standard empirical risk until it converges, resulting in the parameter $\theta^{\star}$. Following this, we compute the influence function and assign appropriate weights to ensure equal True Positive Rate (TPR) and True Negative Rate (TNR) across different groups within the validation set. Unlike previous methods such as those by Ren et al.~\cite{ren2020notall} and Teso et al.~\cite{teso2021interactive}, which calculate the influence function dynamically, our approach computes it only after the model has converged. This strategy not only reduces the computational time required for the influence function but also yields a more precise estimation. 

\begin{algorithm}[h]
\caption{\fairIF.}
\centering
\begin{algorithmic}
\STATE \textbf{Input:}  training set $\mathcal{D}$, validation set $\mathcal{D}^s$, model $f$ and initial paramter $\theta$, hyperparameters $\lambda$.\\
\STATE \textbf{- Stage one: Balancing Influence} \\  
\STATE \textbf{1.} Train $f_{\theta}$ on $\mathcal{D}$ via ERM until converge to obtain $\theta^{\star}$.
\STATE \textbf{2.} Compute $\sum_{z_i \in \mathcal{D}} \nabla_\theta \ell(z_i, \theta^{\star})$  and $H^{-1}_{\theta^\star}$ with stochastic estimation. 
\STATE \textbf{3.} Compute performance differences $\text{diff} (\mathcal{D}^s, F_{TPR}, \theta^\star)$, $\text{diff} (\mathcal{D}^s, F_{TNR}, \theta^\star)$ over $\mathcal{D}^s$, and averaged gradient 
$\frac{1}{|\mathcal{D}^0|} \sum_{z_j \in \mathcal{D}^0} \nabla_\theta{F_{TNR}}(z_j)$, $\frac{1}{|\mathcal{D}^1|} \sum_{z_{j'} \in \mathcal{D}^1} \nabla_\theta{F_{TNR}}(z_{j'})$.\\
\STATE \textbf{4.} Obtain the balancing weight vector $\mathbf{\epsilon}^{\star}$ through optimizing the objective, Equ.~\eqref{equ:final_eps_loss}.\\
\STATE \textbf{- Stage two: Reweighting}  \\
\STATE \textbf{5.} Train $f_{\theta}$ on $\mathcal{D}$ with the loss in Equ.~\eqref{equ:final_goal} to obtain $\theta^{\star}_{\bm{\epsilon^{\star}}}$.\\
\textbf{return} Final model $f_{\theta^{\star}_{\bm{\epsilon^{\star}}}}$.
\end{algorithmic}
\label{alg:fairif}
\end{algorithm}

\section{APPENDIX: Computation Details of Influence Function}
\label{apd:IF_computation}
\header{Computation details of Influence Function.}
As shown in Equation~\eqref{equ:F_diff_change}, the estimation of influence score requires the computation of the inverse hessian. The size of hessian matrix is propotional to the number of model parameters, thus directly computing the inversion of a hessian matrix, \ie $\mathbf{H}_{\theta^{\star}}^{-1}$, is prohibitive.
As described in the previous work~\cite{koh2017understanding}, there are two different ways to efficiently compute $\nabla_\theta F(\{z_{j}\}, \theta^\star)^{\top} H^{-1}_{\theta^\star}\nabla_\theta \ell(z_{i}, \theta^\star)$. The first technique, Conjugate Gradients (CG), is a standard transformation of matrix inversion into an optimization problem. But, as an optimization problem, CG is slow for large dataset.
The second method is LiSSA (Linear time Stochastic
Second-Order Algorithm) method~\cite{agarwal2016second}. LiSSA is a stochastic estimation, which only samples a single point per iteration and results in significant speedups.
Besides, LiSSA provides an unbiased estimation of the Hessian-vector product through implicitly computing it with a mini-batch of samples. As demonstrated in the previous works~\cite{basu2020influence}, the stochastic method is efficient and relatively accurate for sample-wise influence estimation. 
In this work, we employ the second method and the computation of Hessian-vector products (HVPs) can be summarized as:
\begin{itemize}[leftmargin=*]
  \item Step 1. Let $v:= \sum_{z_i \in \mathcal{D}} \nabla_\theta l(z_i)$, and initialize the inverse HVP estimation ${\mathbf{H}}_{0, {\theta^{\star}}}^{-1} v=v$.
  \item Step 2. For $i \in\{1,2, \ldots, J\}$, recursively compute the inverse HVP estimation using a batch size $B$ of randomly sampled a data point $z_{i'}$, $\mathbf{H}_{i, {\theta^{\star}}}^{-1} v=v+\left(I-   \nabla_{\theta}^{2} l(z_i)  \right) \mathbf{H}_{i-1, {\theta^{\star}}}^{-1} v$, where $J$ is a sufficiently large integer so that the above quantity converges.
  \item Step 3. Repeat Step 1-2 $T$ times independently, and return the averaged inverse HVP estimations.
\end{itemize}

\section{APPENDIX: Baselines}
\label{apd:baseline_detail}
In the experiments, we employ the following baselines:
\begin{itemize}[leftmargin=*]
    \item \textbf{CFair}~\cite{zhao2019conditional}. This adversarial approach aims to minimize balanced error rates over the target variable and protected attributes to achieve accuracy equality and equal odds.
    \item \textbf{DOMIND}~\cite{wang2020towards}. This is a domain-independent training scheme that learns a shared feature representation with an ensemble of classifiers for different domains.
    \item \textbf{ARL}~\cite{lahoti2020fairness}. An adversarial optimization strategy that uses computationally identifiable errors to improve worst-case performance over unobserved protected groups.
    \item \textbf{FairSMOTE}~\cite{chakraborty2021bias}. A pre-processing technique that balances internal distributions to ensure equal representation in both positive and negative classes based on the sensitive attribute.
    \item \textbf{Influence} \cite{li2022achieving}. A reweighting strategy that uses a linear programming solver to compute the weights which perfectly bridge the fairness gap.
\end{itemize}

\section{APPENDIX: Dataset Description}
\label{apd:dataset_detail}
\textbf{CI-MNIST.} The Correlated and Imbalanced MNIST (CI-MNIST) is firstly proposed by~\cite{reddy2021benchmarking} to evaluate the bias-mitigation approaches in challenging setups and be capable of controlling different dataset configurations. The label $y\in\{0,1\}$ indicates whether image $x$ is odd or even. And the group attribute $s\in\{0,1\}$ denotes the background color is blue or red. The original dataset assumes that there is a clean and balanced set for test. In this work, we make the distribution of train set and test set to be consistent. As described in Section~\ref{sec:syn_exp}, three different types of bias are independently introduced. The data statistics of them are presented in the Table~\ref{tab:CI-MNIST_stat}. We keep the train/valid/test splits as the original setup~\cite{reddy2021benchmarking}.
\begin{table}[h]
\centering
\small
 \setlength{\tabcolsep}{2pt}{
\begin{tabular}{l | ll | ll}
  \bottomrule
& \multicolumn{2}{c|}{\bf Odd (y=0)} & \multicolumn{2}{c}{\bf Even (y=1)}\\
\hline
 &{\bf Blue (s=0)}&{\bf Red (s=1)}  &{\bf Blue (s=0)}&{\bf Red (s=1)}  \\
\hline
Different Group Size & 30245 & 5337 & 29257 & 5161  \\
Group Dist. Shift & 30245 & 5337 & 5163 & 29255 \\
Different Class Size & 17791 & 17791 & 4303 & 4301 \\
\toprule
\end{tabular}}
\caption{CI-MNIST Data Statistics.} 
\label{tab:CI-MNIST_stat}
\end{table}

\header{Adult}. Each instance in the Adult dataset\cite{asuncion2007uci}  describes an adult with 114 attributes, e.g., gender, education level, age, etc, from the 1994 US Census. We use gender as the sensitive attribute ($s=0$ for female and $s=1$ for male), and the task is to predict whether his/her income is larger than or equal to 50K/year. The data statistics are presented in Table \ref{tab:adult_stat}. We use the train/valid/test splits from the commonly used API aif360 \cite{bellamy2018ai}.

\header{German}. The task in the German dataset\cite{asuncion2007uci} is to classify people as having good or bad credit risks by features related to the economical situation, with gender as the sensitive attribute restricted to female (s=1) and male (s=1). The data statistics are presented in Table \ref{tab:adult_stat}.

\header{COMPAS}. The task in the COMPAS\cite{dieterich2016compas} dataset is to predict recidivism from someone’s criminal history, jail and prison time, demographics, and COMPAS risk scores, with race as the protected sensitive attribute restricted to black (s=0) and white defendants (s=1). The data statistics are also presented in Table \ref{tab:adult_stat}.
\begin{table}[h]
\centering
\small
 \setlength{\tabcolsep}{2pt}{
\begin{tabular}{l | ll | ll}
  \bottomrule
& \multicolumn{2}{c|}{\bf Negative (y=0)} & \multicolumn{2}{c}{\bf Positive (y=1)}\\
\hline
 &{\bf s=0 }&{\bf s=1}  &{\bf s=0}&{\bf s=1}  \\
\hline
Adult & 13026 & 20988 & 1669 & 9539  \\
German & 201 & 499 & 109 & 191\\
COMPAS & 1514 & 1281 & 1661 & 822\\
\toprule
\end{tabular}}
\caption{Tabular Data Statistics.} 
\label{tab:adult_stat}
\end{table}

\header{CelebA.} The CelebA celebrity face dataset is proposed by~\cite{liu2015deep}. Follow the task setup of~\cite{sagawa2019distributionally} in which the label $y$ is set to be the Blond Hair attribute, and the spurious attribute $s$ is set to be the Male attribute: being female spurious correlates with having blond hair. The minority groups are (blond, male) and the majority groups are (blond, female).  We use the standard train/valid/test splits from \cite{sagawa2019distributionally} in the main experiment (Section~\ref{sec:real_exp}).
\begin{table}[h]
\centering
\small
 \setlength{\tabcolsep}{2pt}{
\begin{tabular}{l | ll | ll}
  \bottomrule
& \multicolumn{2}{c|}{\bf Not Blond Hair (y=0)} & \multicolumn{2}{c}{\bf Blond Hair (y=1)}\\
\hline
 &{\bf Female (s=0)}&{\bf Male (s=1)}  &{\bf Female (s=0)}&{\bf Male (s=1)}  \\
\hline
CelebA & 89931 & 28234 & 82685 & 1749  \\
\toprule
\end{tabular}}
\caption{CelebA Data Statistics.} 
\label{tab:CelebA_stat}
\end{table}

\header{FairFace.} FairFace is proposed by~\cite{karkkainen2019fairface}. The face image dataset is balanced on race, gender and age. In our work, we take the gender prediction as the task and denote the $y=1$ as {\it Male} and  $y=0$ as {\it Female}. And the sensitive attribute $s$ s set to be  the race, where $s=0$ denotes {\it Black} and $s=1$ denotes the {\it White}. We keep the train/valid/test splits as the original setup~\cite{karkkainen2019fairface}.
\begin{table}[h]
\centering
\small
 \setlength{\tabcolsep}{2pt}{
\begin{tabular}{l | ll | ll}
\bottomrule
& \multicolumn{2}{c|}{\bf Female (y=0)} & \multicolumn{2}{c}{\bf Male (y=1)}\\
\hline
 &{\bf Black (s=0)}&{\bf White (s=1)}  &{\bf Black (s=0)}&{\bf White (s=1)}  \\
\hline
FairFace & 6894 & 6895 & 8789 & 9823  \\
\toprule
\end{tabular}}
\caption{FairFace Data Statistics.} 
\label{tab:FairFace_stat}
\end{table}\\

\section{APPENDIX: Implementation Details}
\label{apd:implementataion}
In this section, we elucidate the architectures and hyperparameters chosen for each methodology.

\noindent For the three variations of CI-MNIST:
\begin{itemize}[leftmargin=*]
\item {MLP}: Employs a single hidden layer with a dimension of 64.
\item {CNN}: Comprises two convolutional layers, each with filters measuring $5 \times 5$.
\item {LeNet}: Features three convolutional layers.
\end{itemize}
\noindent Across these datasets, we maintained a consistent learning rate of 0.0002, set $\lambda$ to 0.1, and undertook training over 500 epochs.

\noindent For the datasets Adult, German, and COMPAS:
\begin{itemize}[leftmargin=*]
\item We set the logistic regression's learning rate to 0.001 and designated the training epoch count as 250.
\end{itemize}
\noindent For the CelebA and FairFace datasets:
\begin{itemize}[leftmargin=*]
\item We adopted the PyTorch~\cite{paszke2017automatic} versions of ResNet-18, 34, and 50 ~\cite{he2016deep}.
\item Guided by hyperparameter recommendations from ~\cite{liu2021just}, we locked the learning rate at 0.0002 without any learning rate scheduling, set $\lambda$ to 0.1, and capped training epochs at 50. The final layer's hidden dimension was set to 128.
\end{itemize}
\noindent For \fairIF's second stage, we mirrored the training configuration of the first stage. Notably, in Section~\ref{sec:pretrain_exp}, our models begin with weights pretrained on ImageNet.All our experiments were executed using four Tesla V100 SXM2 GPUs, supported by a 12-core CPU operating at 2.2GHz.

\section{APPENDIX: Size of the Validation set}
\label{apd:valid_size_exp}
In Sections~\ref{sec:syn_exp} and~\ref{sec:real_exp}, we employed standard validation sets for each dataset, leveraging their cost-effectiveness due to their size being 5-10 times smaller than training sets. To explore if \fairIF\ could enhance performance with even smaller validation sets—thus reducing group annotation costs—we tested it on the FairFace and CelebA datasets with validation set sizes of 100\%, 50\%, 25\%, and 10\%. Adjusting sample weights and tuning \fairIF\ based on them, Table~\ref{tab:validation} indicates that the fairness-utility trade-off of \fairIF\ is optimal with 50\% or full validation sets, given the refined influence estimations from more group data. The extremely small validation (10\% of original validation set) led to a dip in accuracy, emphasizing the importance of a reasonable validation set for effective parameter tuning.
\begin{table}[!h]
\centering
\caption{Effect of Validation Data on~\fairIF. The Orig. row indicates the performance of the model trained with ERM.} 
\label{tab:valid_set_size}
\vspace{-0.1cm}
\scalebox{0.87}{
 \setlength{\tabcolsep}{3pt}{
\begin{tabular}{l | llll | llll}
  \bottomrule
& \multicolumn{4}{c|}{\bf FairFace} & \multicolumn{4}{c}{\bf CelebA}\\
\hline
 &{\bf Acc}(\%)&{\bf AD}(\%) &{\bf AOD} &{\bf EOD} &{\bf Acc}(\%)&{\bf AD}(\%) &{\bf AOD} &{\bf EOD} \\
\hline
Orig. & 86.68 & 7.23 &  0.0721& 0.0610  & 95.41 & 4.54 & 0.2485 & 0.4975 \\
\hline
Full & {86.63} & {4.14} & {0.0048} & {0.0105} & {95.37} & {3.81} & {0.1220} & {0.3145} \\
50\% & {86.17} & {5.01} & {0.0073} & {0.0152} & {95.59} & {4.14} & {0.1491} & {0.3981} \\
25\% & 86.91 & 5.63 & 0.0089 & 0.0132 & 95.31 & 4.85 & 0.1873 & 0.3748  \\
10\% & 85.49 & 6.44 & 0.0149 & 0.0298 & 93.17 & 5.19 & 0.2219 & 0.4753 \\
\toprule
\end{tabular}}
}
\vspace{-1mm}
\label{tab:validation}
\end{table}